%% file: main.tex
\documentclass[letterpaper]{article} 
\usepackage{aaai2026}  
\usepackage{times}  
\usepackage{helvet}  
\usepackage{courier}  
\usepackage[hyphens]{url}  
\usepackage{graphicx} 
\usepackage{natbib}  
\usepackage{caption} 
\usepackage[utf8]{inputenc} 
\usepackage{booktabs}       
\usepackage{amsfonts}       
\usepackage{nicefrac}       
\usepackage{microtype}      
\usepackage[dvipsnames]{xcolor}         
\usepackage{graphicx}         
\usepackage{amsmath}
\usepackage[linesnumbered,ruled,vlined]{algorithm2e}
\usepackage{makecell}
\usepackage{pifont}
\newcommand{\cmark}{\ding{51}}%
\newcommand{\xmark}{\ding{55}}%

\definecolor{mygreen}{HTML}{97d077}

\title{Autonomous Vehicle Path Planning\\
by Searching With Differentiable Simulation
}

\author{
    Asen Nachkov\textsuperscript{\rm 1}, Jan-Nico Zaech\textsuperscript{\rm 1}, Danda Pani Paudel\textsuperscript{\rm 1}, Xi Wang\textsuperscript{\rm 2}, Luc Van Gool\textsuperscript{\rm 1} 
}
\affiliations{
    \textsuperscript{\rm 1}INSAIT, Sofia University “St. Kliment Ohridski”, Sofia, Bulgaria \\
    \textsuperscript{\rm 2}ETH Zurich, Zurich, Switzerland \\
}

\begin{document}

\maketitle

\input{sections/0_abstract}
\input{sections/1_intro}
\input{sections/2_related_work}
\input{sections/3_method}

\input{sections/4_experiments}

\input{sections/5_conclusion}

\section*{Acknowledgements}
This research was partially funded
by the Ministry of Education and Science of Bulgaria (support for INSAIT, part of the Bulgarian National Roadmap
for Research Infrastructure).

\bibliography{aaai2026}


\end{document}

%% file: sections/0_abstract.tex
\begin{abstract}
Planning allows an agent to safely refine its actions before executing them in the real world. In autonomous driving, this is crucial to avoid collisions and navigate in complex, dense traffic scenarios. One way to plan is to search for the best action sequence. However, this is challenging when all necessary components -- policy, next-state predictor, and critic -- have to be learned. Here we propose Differentiable Simulation for Search (DSS), a framework that leverages the differentiable simulator Waymax as both a next state predictor and a critic. It relies on the simulator's hardcoded dynamics, making state predictions highly accurate, while utilizing the simulator's differentiability to effectively search across action sequences. Our DSS agent optimizes its actions using gradient descent over imagined future trajectories. We show experimentally that DSS -- the combination of planning gradients and stochastic search -- significantly improves tracking and path planning accuracy compared to sequence prediction, imitation learning, model-free RL, and other planning methods.
\end{abstract}

%% file: sections/1_intro.tex
\begin{figure}[h]
    \centering
    \includegraphics[width=1\columnwidth]{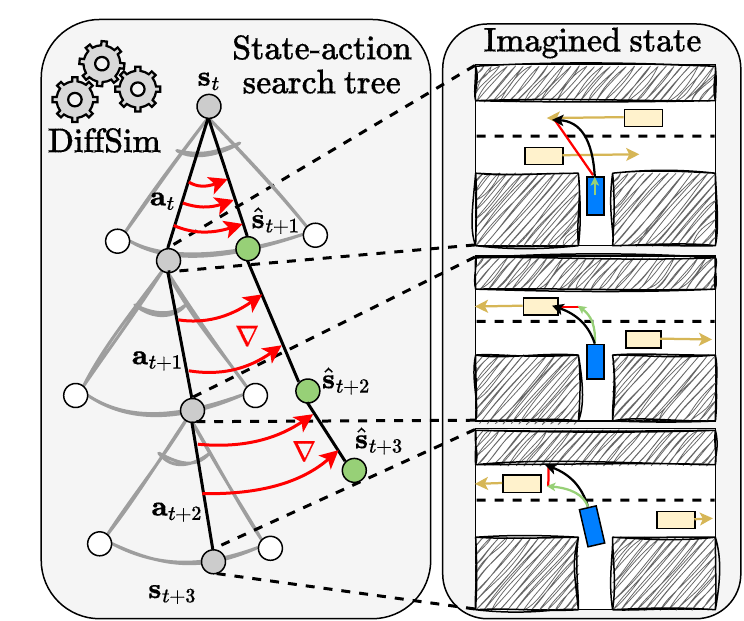}
    \captionsetup{belowskip=-0.35cm, aboveskip=0.1cm}
    \caption{\textbf{Differentiable simulation at test time}. To select the current action, the agent uses a differentiable simulator to imagine a future trajectory ($\mathbf{s}_t, ..., \mathbf{s}_{t+3}$, \textcolor{gray}{gray} circles), sampled from a distribution of possible states (arcs, white circles). The imagined trajectory is \textcolor{red}{refined} using gradient descent towards an optimal one ($\hat{\mathbf{s}}_{t+1}, ..., \hat{\mathbf{s}}_{t+3}$, \textcolor{ForestGreen}{green} circles).}
    \label{fig: aaai_teaser}
\end{figure}

\section{Introduction}
\label{sec: introduction}

When a human driver notices an approaching vehicle in the rear-view mirror, they expect to see the overtaking vehicle in front of them in the next several seconds. This intuitive subconscious anticipation is a form of world modeling and enables planning -- the process of selecting the right actions by predicting and assessing their likely effects. For a computational driving agent, planning is also crucial in order to drive safely and reliably across diverse scenes and conditions. 

One way to perform planning is to \emph{search} for the best action sequence across multiple candidates. The planning agent imagines a number of them, uses its world model to estimate their effects, rates them according to preferences and task constraints, and selects the best one. This is intuitive, yet practical questions are still plentiful. For example, should the agent consider more trajectories or focus and refine a few of them? In this context, we demonstrate that \emph{differentiable simulation} is well-suited for this search problem and enables very efficient planning, as we show below.

Generally, developing accurate search capabilities is not easy. Intuitively, one needs three modules: a policy to suggest actions, a state predictor to predict their effect, and a critic to score them. All modules need to be accurate so the agent can both propose good actions and recognize them as being good. Previous works have attained impressive results in controlled symbolic environments where only some of these modules have to be \emph{learned} \cite{silver2016mastering, moura2021lean}. In fact, the general insight is that environments where we can obtain such components nearly optimally are more suitable for successful planning. This is because a hardcoded realistic simulator used as a state predictor, or a symbolic engine used as a critic, is almost always more accurate than their learned counterparts, facilitating searching.

In the field of autonomous vehicles, test-time search has been underexplored, likely due to challenges like continuous action spaces, diverse realistic scenes, and limited data scale. To search efficiently, we require (i) the next-state predictor and critic to be as accurate as possible, so the agent's imagination faithfully represents a realistic probable future, and (ii) the relationship between the agent's actions and the imagined outcome to be ``easy-to-model", so that a small change in the actions leads to a small change in the imagined outcome. To satisfy these we use the differentiable simulator Waymax \cite{gulino2024waymax} as both a next-state predictor and a critic. Its hardcoded dynamics are realistic, making any sampled Monte Carlo trajectory highly informative. By being differentiable, we can backpropagate through it and any error in the imagined outcome induces a proportional error in the agent's imagined actions. Overall, while a simulator itself provides only evaluative feedback, like in classic reinforcement learning \cite{sutton1998reinforcement}, a differentiable simulator, as used by us, provides instructive feedback through the loss gradients, pointing to the direction of maximal increase and allowing the agent to more efficiently search through the possible action sequences.

Our approach to test-time planning, shown in Fig. \ref{fig: aaai_teaser}, is called \textbf{Differentiable Simulation for Search} (DSS). The agent imagines a future trajectory in which it acts according to a trained policy. The trajectory is \emph{virtual} and is obtained in an autoregressive manner (step by step) using the simulator. Subsequently, the simulator rates the trajectory, based on for example whether there are any collisions or offroad events, and a planning loss is computed. The gradients of this loss propagate through the differentiable dynamics in a manner similar to backpropagation-through-time (BPTT) and reach the ego-vehicle's actions, updating them towards an optimum. The optimized actions can then be executed in the real world, thereby forming a real \emph{physical} trajectory.

Our DSS approach is designed for single-agent control.
Within the virtual future other agents should evolve according
to how the ego-vehicle imagines them. Thus, to obtain their future expected locations, the ego-vehicle needs to perform multiagent control. From its own perspective this is natural, given that everything is controllable within one’s own imagination. From the perspective
of Waymax, however, it requires controlling all agents using
a learned policy. Waymax works by replaying the historical
real-world scenarios from the Waymo Open Motion Dataset
(WOMD) \cite{ettinger2021large}. Those agents not controlled
by a learned, user-provided policy are evolved according to their historic motion, which the ego-vehicle is not supposed to see, creating a potential data leakage. Hence, within the virtual future, all agents have to be controlled.

To run such a planner in a real vehicle, one would need a sensing stack with (i) state \emph{estimation} software that initializes the simulator state from real sensory data (surrounding cameras, LiDAR, IMUs), keeps track of appearing and disappearing objects, receives navigation, and (ii) state \emph{simulation} software, which steps the dynamics forward and generates the virtual trajectories. While Waymax only provides state simulation, we sidestep the need to build state estimation ourselves by adopting the WOMD scenarios and their 9 second episodes, allowing us to study the planning task in isolation.

The Waymax simulator provides differentiable dynamics mapping state-action pairs to next states, which in DSS become the next-state predictor. It also provides calculations like collision and offroad detection, which in our framework  become the critic and have to be used in the planning loss function. Yet, they are boolean and non-differentiable. To be able to extract instructive feedback from them, they have to be differentiable. Thus, we propose \textbf{Classifier-Guided Action Selection} -- a simple approach that uses a small learnable classifier, implemented as a neural network, to detect any undesirable non-differentiable events, such as offroad or collisions. Similar to how classifier-guided diffusion is used to generate images \cite{dhariwal2021diffusion}, this technique is used to steer actions away from collision and offroad events, according to their likelihoods from the classifier.

 This design renders our DSS framework efficient due to the use of a differentiable simulator, realistic owing to the planning, and inherently interpretable.  It guarantees that any actions executed are chosen based on per-scenario optimization, going beyond reactive decision-making. It achieves 16 times lower displacement error than a reactive baseline at tracking, is 2 times better at realistic path planning, and obtains almost 2 times better collision rates compared to state-of-the-art trackers. Overall, our contributions are threefold:
\begin{enumerate}
    \item We propose DSS -- a framework that allows searching for the best action-sequence within a continuous action differentiable simulator at test time, in Sec. \ref{subsec: dss}.
    \item We propose Classifier-Guided Action Selection as a simple way to approximate non-differentiable events and to plan against them, in Sec. \ref{subsec: loss_fn}.
    \item We implement and evaluate the framework for both tracking and path planning, showing its benefits compared to state-of-the-art methods and baselines.
\end{enumerate}

%% file: sections/2_related_work.tex
\section{Related Work \& Context}
\label{sec: related_work}

Our work stands at the intersection of differentiable simulation (DiffSim) and classic RL planning methods.

\textbf{DiffSim for driving.} Differentiable coupling of different modules is not new within autonomous vehicles. TrAAD \cite{zheng2022traffic} and TrafficSim \cite{suo2021trafficsim} have modeled traffic interactions with differentiable ODE-based car-following dynamics or an entirely learned multi-agent model, respectively. DIPP \cite{huang2023differentiable} integrates a differentiable planner, allowing joint optimization of motion prediction and planning objectives. Compared to this, our DSS framework updates the trajectory using gradient descent instead of solving a full optimization problem, and can refine the actions based on multiple likely futures. DiffStack \cite{karkus2023diffstack} composes a fully differentiable AV stack where gradients flow through learned modules like differentiable MPC. Recently, APG \cite{nachkov2024autonomous} has focused on differentiable dynamics, as opposed to author-designed differentiable modules. There, DiffSim makes the policy rollout differentiable, allowing end-to-end training, on which also our framework relies.

\textbf{DiffSim in robotics}. Within robotics, differentiable simulation is rapidly becoming popular. It allows estimating physical object properties from simulation \cite{geilinger2020add} and training robotic policies \cite{lutter2021differentiable, toussaint2018differentiable, qiao2020scalable, holl2020learning}. Differentiable contact models have been used for object manipulation \cite{li2023dexdeform, xu2021end, xu2023efficient, lin2022diffskill}. APG has found applications in aerial navigation with fixed-wing drones \cite{wiedemann2023training}, quadruped locomotion \cite{song2024learning}, and for quadrotor control from visual features \cite{heeg2024learning}. However, in contrast to our approach, in all these applications DiffSim is only used during training, and the policy is simply rolled out at test time.

\begin{table}[t]
\centering
\begin{tabular}{lllll}
\toprule[1.5pt]
\textbf{Method} & \textbf{Policy} & \makecell[tl]{\textbf{Next}\\\textbf{state}} & \textbf{Critic} & \makecell[tl]{\textbf{Action}\\\textbf{selection}} \\
\midrule[1pt]
PG   & L & ---  & --- & Reactive \\
APG  & L & ---   & ---   & Reactive \\
Dyna & L & L & L  & Reactive \\
MPC  & ---   & L/S     & L  & MPC \\
AWM-MPC & L   & L     & L  & MPC \\
AlphaGo & L & S  & L & MCTS  \\
MuZero  & L & L  & L  & MCTS \\
AlphaGeometry& L & SE   & L  & BS \\
AlphaProof   & L & PS      & L  & MCTS \\
AlphaTensor  & L & S  & L & MCTS  \\
LLM reasoning & L & --- & L & BS \\
DSS (ours)  & L & S & S + L & MPC + $\nabla$  \\
\bottomrule[1.5pt]
\end{tabular}
\captionsetup{belowskip=-0.4cm}
\caption{\textbf{Relevant method characteristics}. DSS (ours) is the only one that uses differentiable simulation at test time. L = learned neural net, S = simulator, SE = symbolic engine, PS = proof system, MCTS = Monte Carlo tree search, BS = beam search, MPC = model predictive control, $\nabla$ = gradients.}
\label{table: method_comparison}
\end{table}

\textbf{Planning.} In RL, many promising models have used planning in diverse environments, attaining strong results. We list them in Table \ref{table: method_comparison}, highlighting their \emph{main} differences. Methods like policy gradients (PG) \cite{sutton1999policy} or DQN \cite{mnih2015human} are reactive in nature since they act by following a learned explicit or implicit policy. APG follows them but uses DiffSim at training time. Dyna-style algorithms \cite{sutton1991dyna, hafner2019dream} use a world model to train on imagined transitions but are reactive at test time. MPC uses a world model to predict trajectories, rate them, and select the best one. Variants based on DiffSim have also been developed \cite{nachkov2025dream}. For board games, AlphaGo \cite{silver2016mastering} and MuZero \cite{schrittwieser2020mastering} use sophisticated planning in discrete action spaces, while methods like AlphaGeometry \cite{trinh2024solving}, AlphaProof \cite{AlphaProof2024}, and AlphaTensor \cite{fawzi2022discovering} rely on symbolic engines or proof systems to facilitate the search. Recently, LLMs have been used as policies and been combined with test-time search strategies such as Best-of-$N$ or beam search \cite{snell2024scaling}. There, a process reward model acts as a learned critic to score answers \cite{luo2024improve}. Compared to all these methods, our DSS planning framework is the only one that uses a differentiable simulator at test time to search for the right action. Like them, we use a learned policy that captures the right statistical patterns from the scenarios at training time.

\textbf{State-of-the-art in Waymax.} Within the Waymax setting, there are hardly any baselines that use planning. We compare against behavior cloning \cite{gulino2024waymax} and APG approaches, sequence prediction approaches such as Wayformer \cite{nayakanti2023wayformer}, offline RL methods like Decision Transformer (DT) \cite{chen2021decision} and more recent Waymax state-of-the-art RL baselines like EasyChauffeur \cite{xiao2024easychauffeur} and PiDT \cite{zhou2025physicsinformedimitativereinforcementlearning}. None of these are exact alternatives to our planning framework because they are designed to rely on privileged information. For example, EasyChaffeur uses the full historic trajectory as a route to condition the agent (more akin to path tracking, rather than autonomous navigation), whereas our setup is more realistic and only uses the last $(x, y)$ waypoint to indicate a final destination. Additionally, previous methods have different training objectives, often aiming to reproduce historical expert actions, whereas our training aims to reproduce historical expert states -- a subtle nuance.

%% file: sections/3_method.tex
\section{Method}
\label{sec: method}

\textbf{Notation.} We represent the set of all states as $\mathcal{S}$ and that of the actions as $\mathcal{A}$. The simulator is abstracted as a pure, stateless differentiable function, $\text{Sim}: \mathcal{S} \times \mathcal{A} \rightarrow \mathcal{S}$, that maps state-action pairs to next states, $\text{Sim}(\mathbf{s}_t, \mathbf{a}_t) \mapsto \mathbf{s}_{t + 1}$. A trajectory is a sequence of state-action pairs $(\mathbf{s}_0, \mathbf{a}_0, \mathbf{s}_1, \mathbf{a}_1, ..., \mathbf{s}_T)$. We can extract agent locations $(\mathbf{x}_0, \mathbf{x}_1, ..., \mathbf{x}_T)$ and action sequences $(\mathbf{a}_0, ..., \mathbf{a}_{T-1})$ from it. We denote an  action of the $e$go-vehicle as $\mathbf{a}_t^e$, and an action of all $o$ther agents as $\mathbf{a}_t^{o}$. Actions are vector-valued with a dimension $A$ and represent acceleration and steering in our setting.

\textbf{DSS agent}. Our DSS framework requires a learned stochastic policy to model agent behavior. Its training is described in Sec. \ref{subsec: training}, and our approach to planning at test time in Sec. \ref{subsec: dss}. Classifier-Guided Action Selection, in Sec. \ref{subsec: loss_fn}, allows us to model non-differentiable events when planning.

\subsection{Training -- Analytic Policy Gradients}
\label{subsec: training}

We need to learn a stochastic policy $\pi_\theta$ for producing the reactive behavior of the agents. We train it using Analytic Policy Gradients (APG) \cite{nachkov2024autonomous} to learn a realistic action distribution from historical expert driver trajectories. Specifically, we train on the WOMD scenarios within Waymax. In each scenario the agent performs a rollout, after which the full obtained trajectory $(\mathbf{s}_0, ..., \mathbf{s}_T)$ is supervised with the expert human driver one, $(\hat{\mathbf{s}}_0, ..., \hat{\mathbf{s}}_T)$. Gradients flow through the dynamics, similar to BPTT:
~
\begin{equation}
\begin{aligned}
\min_\theta \ \mathcal{L}_\text{train} &= \lVert \mathbf{s}_t - \hat{\mathbf{s}}_t\rVert_2^2 \\
\text{ with } \mathbf{s}_t &= \text{Sim}(\mathbf{s}_{t-1}, \pi_\theta(\mathbf{s}_{t-1})).
\end{aligned}
\end{equation}

\textbf{Action selection}. The policy $\pi_\theta$ is stochastic and is parametrized as a Gaussian mixture with six components. To encourage action multimodality during training, actions for the rollout are selected by sampling not from the entire Gaussian mixture, but only from that one component that will bring the ego-vehicle closest to the next expert state \cite{nayakanti2023wayformer}. We use the simulator to find that component efficiently (details in the suppl. materials). The error signals during backpropagation reach only this component, instead of all of them. This allows the policy to sample diverse actions, which is beneficial for searching at test time.

\begin{algorithm}[t]
\DontPrintSemicolon
\KwIn{Initial state $\mathbf{s}_0$, policy $\pi_\theta$, simulator Sim, loss function $\ell$, imagination horizon $T$, number of rollouts $K$, discount factor $\gamma$, temperature $\tau$, step size $\eta$}
\KwOut{Ego action $\mathbf{a}_0$ to apply at initial state $\mathbf{s}_0$}

\vspace{4pt}
\For{$k \gets 0$ \KwTo $K - 1$}{
    Extract agent states $\mathbf{x}_0^{e,k}, \mathbf{x}_0^{o,k} \gets \mathbf{s}_0$\;
    \For{$t \gets 0$ \KwTo $T - 1$}{
        $\mathbf{a}_t^{e,k}, \mathbf{a}_t^{o,k} \gets \pi_\theta(\mathbf{x}_{\leq t}^{e,k}, \mathbf{x}_{\leq t}^{o,k})$\;
        $\mathbf{x}_{t+1}^{e,k}, \mathbf{x}_{t+1}^{o,k} \gets \text{Sim}(\mathbf{x}_t^{e,k}, \mathbf{x}_t^{o,k}, \mathbf{a}_t^{e, k}, \mathbf{a}_t^{o,k})$\;
    }
    Compute rollout losses $\ell^k = \ell(\mathbf{x}_{1:T}^{e,k})$ for all $k$\;
    Compute ego action gradients $\mathbf{g}_0^k = \partial \ell^k / \partial \mathbf{a}_0^{e,k}$\;
}

Compute weights: $w_k = \big(\exp(- \ell^{k} / \tau)\big)/\big(\sum_{j=0}^{K-1} \exp(- \ell^{j} / \tau)\big)$\;

\Return $\sum_{k=0}^{K-1} w_k \big( \mathbf{a}_0^{e,k} - \eta \mathbf{g}^k_0 \big)$
\caption{DSS}
\label{alg: main_alg}
\end{algorithm}

\textbf{Recurrent architecture}. Since the policy is recurrent, its hidden state encapsulates the entire history of observations, represented as $\pi_\theta(\mathbf{a}_t | \mathbf{s}_{\le t})$. For computational efficiency during training we only control the ego-vehicle, while the other agents' states evolve according to their historic motion. However, at test time the policy could be used to control also the other agents. By extracting state observations from the perspective of all agents we can compute all actions in parallel. We overwrite the notation as $\mathbf{a}_t^e, \mathbf{a}_t^o = \pi_\theta(\mathbf{x}_{\le t}^e, \mathbf{x}_{\le t}^o)$, where $\mathbf{x}_t^e$ and $\mathbf{x}_t^o$ indicate the $e$go and $o$ther agents' positions at time $t$. Full implementation details are in the suppl. materials.

\subsection{Testing -- Differentiable Simulation for Search}
\label{subsec: dss}

\begin{figure*}[t]
    \centering
    \includegraphics[width=1\textwidth]{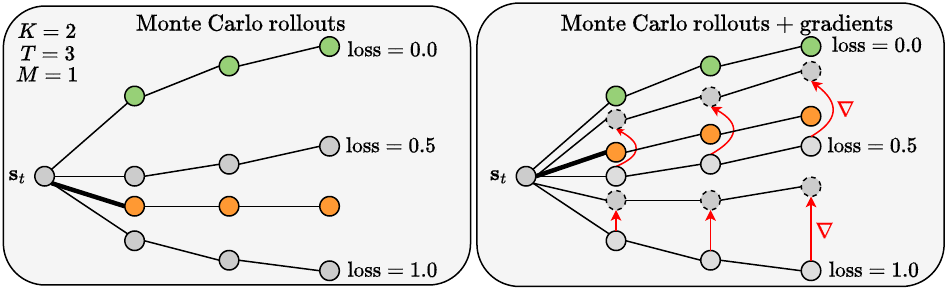}
    \captionsetup{belowskip=-0.35cm, aboveskip=0.1cm}
    \caption{\textbf{Gradient descent in the ego-agent's imagined future}. Without gradients (left), searching involves sampling $K$ trajectories (\textcolor{gray}{gray}) of length $T$, scoring them, and aggregating their first $M$ actions (\textbf{bold} black line). The trajectory from the resulting actions is shown in \textcolor{orange}{orange}. With gradients, each rolled out trajectory is first updated towards an optimum (\textcolor{ForestGreen}{green}), as judged by the planning loss. The executed trajectory from the aggregated actions more closely aligns with this optimal trajectory.}
    \label{fig: dss}
\end{figure*}

Our planning algorithm at test time, called Differentiable Simulation for Search (DSS), is shown in Algorithm \ref{alg: main_alg}. To select the current action, the ego-agent \emph{imagines} $K$ future trajectories, each of length $T$ steps. They are generated autoregressively by using the trained policy $\pi_\theta$ to compute actions for both the ego- and the other agents, while the simulator is used to compute their next states, in lines \texttt{3-5}. Having obtained the $K$ trajectories, we compute a loss function over the ego positions, line \texttt{6}. With DiffSim, in lines \texttt{7} and \texttt{9} we can compute the gradients of the loss with respect to the first ego-action and perform a single gradient descent step to improve it. Since the algorithm is based on sampled imaginary rollouts, the final selected action is a weighted average of the optimized first actions from these rollouts, where actions in trajectories with lower losses have higher weights. Fig. \ref{fig: dss} shows how the gradients improve the trajectory.

\textbf{Flexibility}. Alg. \ref{alg: main_alg} encompasses a full set of possible behaviors for how the agent can plan its action at test time (see Fig. \ref{fig: ablation_tree}). The setting $K=1, T=1, \eta = 0$ represents a \textbf{reactive agent} that drives by relying on the trained policy $\pi_\theta$. If $K=1$ but $\eta > 0$, the agent uses the differentiability of the simulator to optimize its actions, as the gradient step size $\eta$ is positive, but does not use the simulator to perform Monte Carlo search. We call this setting \textbf{reactive with gradients}. If $K > 1$ and $\eta = 0$ the agent uses multiple Monte Carlo rollouts to search for the right actions but does not use gradient descent to optimize them. This setting is called \textbf{simulator as a critic}, because the simulator computes trajectories and losses, but its differentiability is not used. Finally, our full proposed setting \textbf{differentiable simulator as a critic} is enabled when gradients are used to optimize the actions, i.e. $\eta > 0$, and multiple rollouts are used to search, i.e. $K > 1$. 

\begin{algorithm}[t]
\DontPrintSemicolon
\KwIn{Initial state $\mathbf{s}_0$, rollout horizon $T$, rollouts $K$, actions to execute $M$}
\KwOut{Continuous sequence of executed ego actions}
\BlankLine
 $\mathbf{s} \gets \mathbf{s}_0$\;
 \While{$\mathbf{s}$ not final}{
    $\mathbf{a}_{0:M-1} \;\gets\; \text{Plan}\bigl(\mathbf{s}, K, T, M)$ \# Algorithm \ref{alg: main_alg}\;
    \For{$i \gets 0$ \KwTo $M-1$}{
       $\mathbf{s}_{t+1} \gets \text{Sim}(\mathbf{s}_t , \mathbf{a}_t)$\;       
    }
 }
 \caption{Main Control Loop}
 \label{alg: exec_loop}
\end{algorithm}

\textbf{Control loop}. The imagined trajectories in Alg. \ref{alg: main_alg} are virtual -- they represent the future as predicted by the ego-agent's policy. Algorithm \ref{alg: exec_loop} provides the main loop for obtaining a real, physical trajectory, over which the evaluation metrics are calculated. Specifically, instead of planning out only the first action, we optimize and execute the first $M$ imagined actions, after which the ego-vehicle has to re-plan. In line \texttt{5}, only the ego-agent is controlled by the policy's optimized actions. This is in contrast to line \texttt{5} in Alg. \ref{alg: main_alg} where all agents are controlled. The overall effect is that Alg. \ref{alg: main_alg} shows \emph{how the ego-vehicle optimizes its own actions within the virtual, imagined dynamics, which inherently involves multi-agent control in order to imagine the other agents' motions}, while Alg. \ref{alg: exec_loop} is used to obtain the real physical trajectories, where only the ego-vehicle is controlled by the policy $\pi_\theta$.

\textbf{Computational cost}. When re-planning once every $M$ steps, the reaction time to any observation can be up to $M$ steps. Re-planning once every $3$ steps corresponds to a reaction time of at most $0.3$ seconds (at 10 frames per second). The total computational cost, in number of policy calls, is $O(LKTN/M)$ where $L$ is the length of the physical trajectory, and $N$ is the maximum number of actors in the scene. For Waymax, where $L=90$ and $N=128$, and when $M=3, K=8, T=10$, this runs at $4.1$ seconds per scenario on a single RTX3090 GPU. Thus, a full scenario long $90$ timesteps, or $9$ seconds of historical real driving, is processed in $4.1$ seconds -- effectively \emph{real-time}.

\subsection{Classifier-Guided Action Selection}
\label{subsec: loss_fn}

The function $\ell$ in line \texttt{6} in Alg. \ref{alg: main_alg} is the main objective to optimize when planning. It should contain loss terms for any undesirable behavior, while not leaking any privileged information such as future ground-truth locations. The \textbf{offroad} and \textbf{collision} functions in Waymax are of interest, yet are boolean and non-differentiable. Previous works \cite{nachkov2025dream} have used 2D Gaussians to approximate the rotated boxes, which yields a closed-form, differentiable overlap formula, but is limited, as it applies only to collisions.

Inspired by classifier-guidance for image generation, we propose Classifier-Guided Action Selection. We approximate the non-differentiable collision and offroad detection functions with a simple multi-label classifier, $p_\phi$, trained on simulated trajectories. We indicate the collision and offroad events as $c$ and $o$, respectively. During training the simulator computes the ground-truth binary labels $M_c$ and $M_o$ for whether such events are present. Then, in Eqn. \ref{eqn: planning_loss_bce} we formulate the collision and offroad losses -- negative binary cross-entropy -- with which the classifier is trained.
~
\begin{equation}\label{eqn: planning_loss_bce}
\begin{aligned}
        &\ell_t^c = -\Big(M_\text{c}\log p_\phi(c | \mathbf{s}_t) + (1 - M_c)\log\big(1 - p_\phi(c | \mathbf{s}_t)\big)\Big)\\
        &\ell_t^o =  -\Big(M_\text{o}\log p_\phi(o | \mathbf{s}_t) + (1 - M_o)\log\big(1 - p_\phi(o | \mathbf{s}_t)\big)\Big)\\
\end{aligned}
\end{equation}
The full loss to minimize during planning is the average collision and offroad probability across all imagined steps:
~
\begin{equation}\label{eqn: planning_loss}
\begin{aligned}
    \min_{\mathbf{a}_t, ..., \mathbf{a}_{t+T}} \ell = \frac{1}{T}\sum_{t=0}^{T-1}\Big(M_c p_\phi(c|\mathbf{s}_t) + M_o  p_\phi(o | \mathbf{s}_t)\Big).
\end{aligned}
\end{equation}
The gradient $\partial \ell / \partial \mathbf{a}_t^{e}$ goes through the classifier $p_\phi$, whose weights $\phi$ are frozen, through the state $\mathbf{s}_{t+1}$, the differentiable dynamics $\text{Sim}(\mathbf{s}_t, \mathbf{a}_t)$, and reaches the actions $\mathbf{a}_{t}$. In this way, thanks to differentiable simulation, the ego-agent can optimize its actions at test time. Importantly, even if some desirable components are non-differentiable, they can still be optimized through search, e.g. when $\eta=0$ and $K > 1$. Hence, our DSS framework is flexible in handling non-differentiable losses, in which case it simply becomes loss-weighted Monte Carlo search over the sampled trajectories.

%% file: sections/4_experiments.tex
\section{Experiments}
\label{sec: experiments}

In this section we validate our proposed approach experimentally. The two research questions we seek to answer are:
\begin{itemize}
\itemsep-1em 
\item \emph{Is our proposed search procedure beneficial in general?} We assess this in Sec. \ref{subsec: tracking} by adopting a well-behaved planning loss function that models a tracking problem. \\
\item \emph{Is it beneficial specifically for AV path planning?} For this we adopt a realistic, restrictive planning loss function in Sec. \ref{subsec: autonomous_path_experiments} that models a difficult path planning problem.\\
\end{itemize}
\vspace{-5pt}
\textbf{Experimental setup}. The \textbf{inputs} to the ego-agent's policy $\pi_\theta$ include the locations of all agents, the nearest roadgraph points, the traffic lights, the agent's own speed, and the last $(x, y)$ waypoint from the expert trajectory, which is needed to mark the final destination (otherwise, without an intended destination one cannot expect to compare to the expert trajectory). The \textbf{outputs} are actions -- acceleration and steering.

We follow the evaluation protocol in previous works \cite{nachkov2024autonomous} and use the Waymax simulator, which builds over the WOMD scenarios. For each scenario we compute the \textbf{displacement error} (= ADE = $L_2$ distance) of the ego-vehicle's physical trajectory compared to the historic one. The number reported is averaged over the timesteps within the trajectory and over all scenarios in the validation set. Further, we track the \textbf{overlap} and \textbf{offroad rates}. They indicate the proportion of scenarios in which at least one ego collision or offroad event occurs.

\subsection{Evaluating the Search Framework}
\label{subsec: tracking}
Here we evaluate the general DSS framework presented in Sec. \ref{subsec: dss}. The planning loss function, line \texttt{6} in Alg. \ref{alg: main_alg}, is the $L_2$ distance between the simulated and the expert trajectory:
\vspace{-5pt}
\begin{equation}
    \ell = \frac{1}{T}\sum_{t=0}^{T-1}\lVert \mathbf{x}^e_t - \hat{\mathbf{x}}^e_t \rVert_2^2
\vspace{-2pt}
\end{equation}
This setting represents a tracking problem -- given a path of time-dependent waypoints $\hat{\mathbf{x}}_0^e, ..., \hat{\mathbf{x}}_{T-1}^e$, optimize for the actions that follow it. The best action for the next timestep necessarily belongs to the sequence of best actions for all future timesteps. Hence, the agent can afford to be greedy and not plan very far ahead into the future (we can set $T=1$).

\begin{figure}[t]
    \centering
    \includegraphics[width=1\columnwidth]{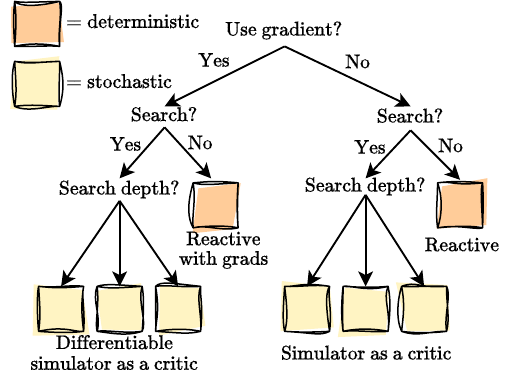}
    \captionsetup{belowskip=-0.15cm, aboveskip=0.1cm}
    \caption{\textbf{Experiment ablation tree}. Depending on the different configuration of whether to search and use gradients, there are four different experimental settings. Our full framework DSS uses both search across multiple trajectories and gradients to optimize the actions across them.}
    \label{fig: ablation_tree}
\end{figure}

\textbf{Evaluations of different settings}. Importantly, we visualize the logical relationships between the different framework settings as an ablation tree, shown in Fig. \ref{fig: ablation_tree}. Depending on whether gradient updates are used ($\eta > 0$), and whether searching is used ($K > 1$), four configurations are available. 

\begin{table}[h]
    \small
    \centering
    \begin{tabular}[width=\textwidth]{ l l | c c c } \toprule[1.5pt]
          $\eta_\text{accel}$ & $\eta_\text{steer}$ & ADE $\downarrow$ & overlap $\downarrow$ & offroad $\downarrow$ \\
         \midrule[1pt]
         0 & 0   & 2.9792 &  0.1369 & 0.0780 \\
         100 & 0.01  & 2.1294 &  0.0816 & 0.0718 \\   
         200 & 0.01  & 1.5282 & 0.0558 & 0.0686  \\
         500 & 0.01 & 0.7193 & 0.0327 & 0.0660 \\
         1000 & 0.01 & \textbf{0.4644} & \textbf{0.0269} & 0.0664 \\
         5000 & 0.01 & 0.7678 & 0.0311 & \textbf{0.0653} \\
         \bottomrule[1.5pt]
    \end{tabular}
    \captionsetup{belowskip=-0.45cm}
    \caption{\textbf{Effect of the gradient step size in reactive settings.} Here we set $T=1$. The best results require a large step size for the acceleration and small one for the steering. \emph{Takeaway}: guiding the reactive policy using the gradient could greatly improve performance, even when not searching.}
    \label{table: step_size_ablations}
\end{table}

The \textbf{reactive} settings are shown in Table \ref{table: step_size_ablations}. Results are deterministic because when $K=1$, the agent only imagines a single trajectory, which we take to be formed by always selecting the mean action (there is no reason to choose otherwise). Since performance is more sensitive to wrong steering than to wrong acceleration, we use different learning rates for the action elements. A high learning rate for the acceleration (1000) is beneficial, yet a too high one (5000) starts causing the update to overshoot the optimal action.

In the \textbf{simulator as a critic} setting we perform search when $K > 1$ without gradient optimization. Table \ref{table: mc_without_grads} shows that results improve as more trajectories are imagined. By imagining different likely future sequences, the agent can find, \emph{by chance}, more accurate action sequences.
\begin{table}[h]
    \small
    \centering
    \begin{tabular}[width=\textwidth]{ l | c c c } \toprule[1.5pt]
         \textbf{Rollouts} $K$ & ADE $\downarrow$ & overlap $\downarrow$ & offroad $\downarrow$ \\
         \midrule[1pt]
         1  & 2.9792 &  0.1369 & 0.0780 \\
         2  & 1.1954 &  0.0381 & 0.0663 \\
         4  & 1.0968 &  \textbf{0.0369} & 0.0653 \\   
         8  & \textbf{1.0710} &  0.0378 & \textbf{0.0646} \\   
         \bottomrule[1.5pt]
    \end{tabular}
    \captionsetup{aboveskip=0.3cm, belowskip=-0.3cm}
    \caption{\textbf{Simulator as predictor and critic.} We sample $K$ trajectories from a stochastic policy with $T=1, \eta=0, \tau=1$. \emph{Takeaway}: using the simulator to score the Monte Carlo trajectories and to search through them yields strong results.}
    \label{table: mc_without_grads}
\end{table}

Finally, Table \ref{table: mc_with_grads} shows the same setting but with gradient updates enabled in addition to the search, this being the full proposed \textbf{differentiable simulator as a critic} setting. The agent can now both find good actions by chance, and further optimize them through the differentiable simulator. By setting the sampling temperature $\tau$ to be low, the agent can select the best action. Note that in some cases selecting the maximum from multiple noisy values, as done for example in DQNs \cite{mnih2015human}, may introduce instability and hurt performance. Here, since the simulator is a perfect critic, we avoid this issue. The standard deviations for the 3 metrics over 5 random seeds are $(0.01, 0.0008, 0.0007)$ and statistical variability plays almost no role in our results.

\begin{table}[h]
    \small
    \centering
    \begin{tabular}[width=\textwidth]{ l | c c c } \toprule[1.5pt]
         \textbf{Model} & ADE $\downarrow$ & overlap $\downarrow$ & offroad $\downarrow$ \\
         \midrule[1pt]
         DecisionTransformer & 8.3200 & 0.0362 & 0.0621 \\
         BC & 3.6000 & 0.1120 & 0.1359 \\
         PiDT & 6.9900 & 0.0186 & 0.0298 \\
         Wayformer & 2.3800 & 0.1068 & 0.0789 \\
         EasyChaffeur-IL & --- & 0.0293 & 0.0280 \\
         EasyChaffeur-PPO & --- & 0.0443 & \textbf{0.0216} \\
         Ours ($K=1$) & 0.4644 & 0.0269 & 0.0664 \\
         Ours ($K=2$) & 0.2509 & 0.0196 & 0.0498 \\
         Ours ($K=4$) & 0.2013 & 0.0168 & 0.0414 \\
         Ours ($K=8$) & \textbf{0.1766} & \textbf{0.0150} & 0.0350 \\
         \bottomrule[1.5pt]
    \end{tabular}
    \captionsetup{aboveskip=0.3cm, belowskip=-0.3cm}
    \caption{\textbf{DSS for trajectory tracking.} Here $T=1, \tau=0.01$ and the learning rate is $\eta = (1000, 0.01)$. \emph{Takeaway}: the search allows the agent to find good actions by chance, while the gradients further refine them.}
    \label{table: mc_with_grads}
\end{table}

\textbf{Discussion.} The results validate the benefits of our DSS procedure. For tracking a log-trajectory the searching and gradient updates improve over the reactive performance by up to $16.9$ times (2.979 $\rightarrow$ 0.176 ADE). This occurs because (i) the Monte Carlo searching against the simulator provides very accurate information about the trajectory's quality, and (ii) the simulator is differentiable and we can update the actions to minimize the loss function. Compared to other methods -- behavior cloning \cite{gulino2024waymax}, Wayformer \cite{nayakanti2023wayformer}, EasyChaffeur \cite{xiao2024easychauffeur}, and PiDT \cite{zhou2025physicsinformedimitativereinforcementlearning} we obtain consistently better ADE and overlap, and competitive offroad rates.

\subsection{Evaluating the Classifier-Guided Actions}
\label{subsec: autonomous_path_experiments}

Having established that the DSS framework is useful, we now turn to evaluating the classifier-guided action selection proposed in Sec. \ref{subsec: loss_fn}. Here the agent only sees a single final $(x, y)$ waypoint that marks its destination and must autonomously plan the intermediate trajectory to it. This represents a significantly more difficult and realistic AV problem setting.

\begin{figure}[t]
    \centering
    \includegraphics[width=0.97\columnwidth]{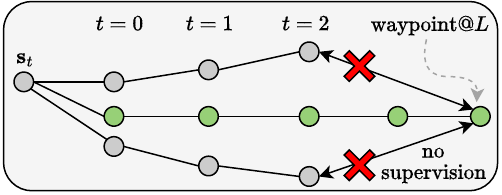}
    \captionsetup{belowskip=-0.35cm, aboveskip=0.1cm}
    \caption{\textbf{Planning loss design}. The planning loss includes collision and offroad events. It does not supervise the last imagined location at time $T$ with the last waypoint at time $L$.}
    \label{fig: conceptual_difficulty}
\end{figure}

\textbf{Problem setting}. A general conceptual challenge is how to incorporate a differentiable planning loss term encouraging the policy to reach the destination indicated by the last waypoint. That waypoint is fixed to the last physical timestep $L$ which \emph{may lie beyond the planning horizon} $T$. This creates a mismatch and prevents us from directly supervising the imagined location at time $T$ with that waypoint beyond $T$. The difficulty arises from the difference between reaching the destination at a \emph{particular} timestep vs reaching it at \emph{any} timestep. To avoid this problem we do not supervise with the waypoint (see Fig. \ref{fig: conceptual_difficulty}). The planning loss function is simply Eqn. \ref{eqn: planning_loss}. The agent is expected to have learned how to time its movement based on the scenarios seen during training.

Collisions and offroad events are sparse and occur only in some steps. By approximating them with a classifier, as described previously, we can minimize the likelihood of such events at test time. Table \ref{table: autonomous_path_planning} shows the exact results. Crucially, \emph{if the classifier is accurate, minimizing their probabilities during the imagination results in selecting those actions that avoid such events in the real physical trajectory}. As before, the search and the gradient updates are beneficial.

\begin{table}[h]
    \small
    \centering
    \begin{tabular}[width=\textwidth]{ l l l | c c c } \toprule[1.5pt]
          $K$ & $T$ & Grad & ADE $\downarrow$ & overlap $\downarrow$ & offroad $\downarrow$ \\
         \midrule[1pt]

         1 & 20 & \xmark  & 2.9872 & 0.1380 & 0.0781 \\
         1 & 20 & \cmark  & 2.9817 & 0.1346 & 0.0737 \\ 
         
         4 & 10 & \xmark  & 1.6016 & 0.0618 & 0.0641 \\
         4 & 10 & \cmark & 1.3861 & 0.0505 & \textbf{0.0638} \\ 
         4 & 20 & \cmark &  1.3235 & 0.0510 & 0.0782 \\ 
         8 & 10 & \cmark & \textbf{1.2759} & 0\textbf{.0442} & 0.0762 \\ 
         \bottomrule[1.5pt]
    \end{tabular}
    \captionsetup{aboveskip=0.3cm, belowskip=-0.3cm}
    \caption{\textbf{Performance when planning only to minimize collisions and offroad.} Gradient step size is set to $\eta = (1000, 0.05)$.  \emph{Takeaway}: searching improves performance ($K=1$ vs $K>1$). A longer planning horizon $T$ (rows \texttt{4-5}) and gradients from the differentiable simulator (\cmark vs \xmark), even though sparse, are still beneficial.}
    \label{table: autonomous_path_planning}
\end{table}

\begin{table}[b]
    \small
    \centering
    \begin{tabular}[width=\textwidth]{ l | c c c } \toprule[1.5pt]
         \textbf{Model} & ADE $\downarrow$ & overlap $\downarrow$ & offroad $\downarrow$ \\
         \midrule[1pt]
         DecisionTransformer & 8.3200 & 0.0362 & 0.0621 \\
         BC & 3.6000 & 0.1120 & 0.1359 \\
         PiDT & 6.9900 & \textbf{0.0186} & 0.0298 \\
         Wayformer & 2.3800 & 0.1068 & 0.0789 \\
         EasyChaffeur-IL & --- & 0.0293 & 0.0280 \\
         EasyChaffeur-PPO & --- & 0.0443 & \textbf{0.0216} \\
         Ours (best configuration) & \textbf{1.2759} & 0.0442 & 0.0762 \\
         \bottomrule[1.5pt]
    \end{tabular}
    \captionsetup{aboveskip=0.3cm, belowskip=-0.3cm}
    \caption{\textbf{Comparison with additional baselines}. \emph{Takeaway}: our planning method outperforms state-of-the-art methods on ADE in a more realistic setting with less route conditioning.}
    \label{table: final_results}
\end{table}

Importantly, even though the agent has not been explicitly trained to minimize overlap and collisions, the test time planning can nonetheless improve these metrics -- ovelap rate improves by 9.37 perc. points (0.138 $\rightarrow$ 0.044), while ADE decreases by more than $2$-fold (2.987 $\rightarrow$ 1.2759). Compared to state-of-the-art methods in Table \ref{table: final_results}, we attain significantly better ADE, which is more important than overlap and offroad alone, because it implicitly contains humanlike pacing, turning, and accelerating. EasyChaffeur and PiDT are directly trained to minimize collision and offroad and obtain better rates there. Unlike EasyChaffeur, our agent only sees the final $(x, y)$ destination and has to decide where to go in the intermediate trajectory up to it. In this more realistic setting we obtain a 5-fold reduction in ADE compared to the other methods (1.27 vs PiDT's 6.99).

\begin{figure}[h!]
    \centering
    \includegraphics[width=1\columnwidth]{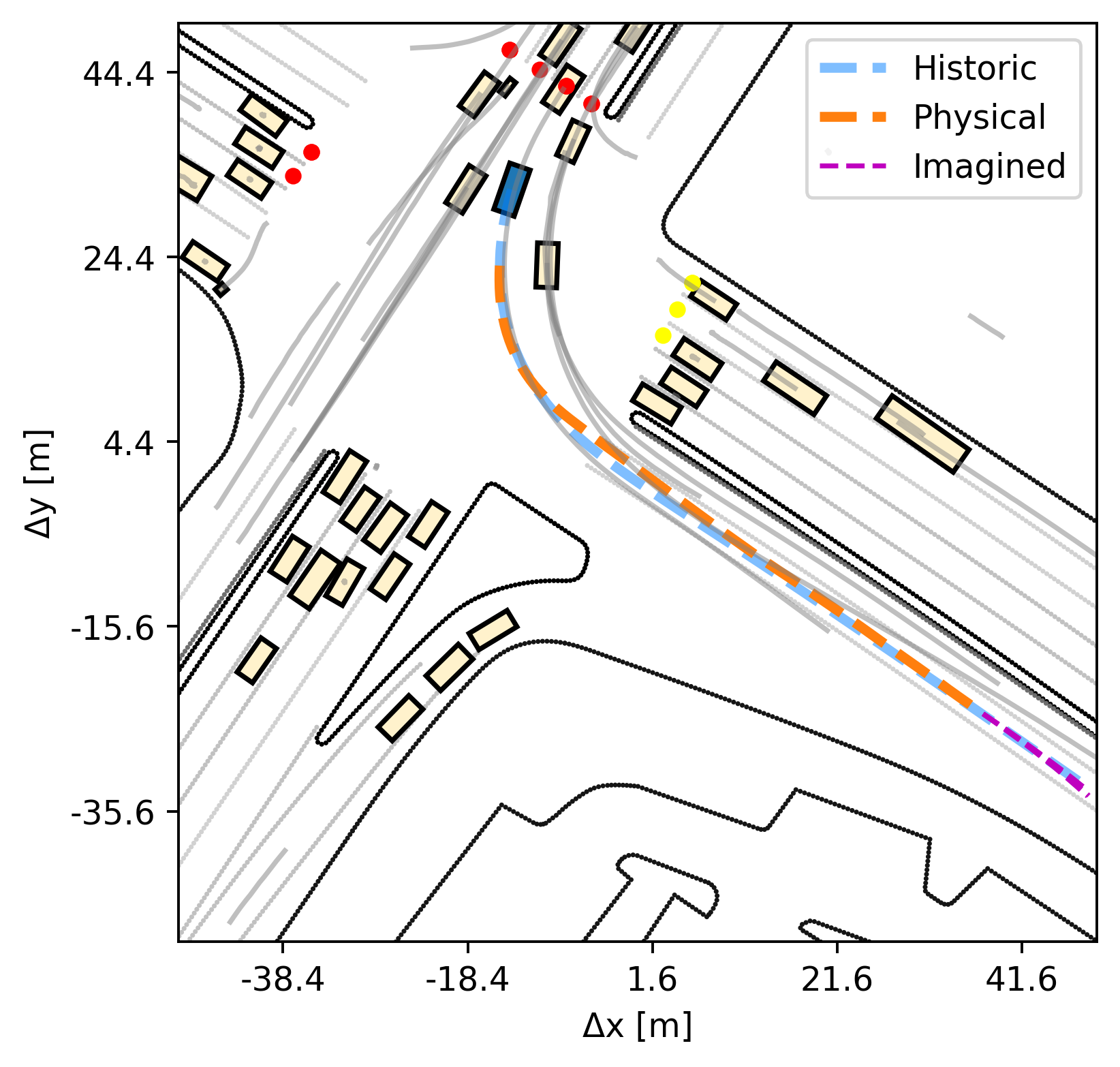}
    \captionsetup{belowskip=-0.25cm, aboveskip=0.0cm}
    \caption{\textbf{Driving by planning}. The ego-vehicle is \textcolor{RoyalBlue}{blue}. All boxes are shown at their initial positions and the \textcolor{gray}{gray} lines indicate their future motion (crossing lines do not imply collisions). \textcolor{red}{Red} dots are red lights. The ego-agent accurately navigates the intersection by periodically planning out its actions through imagination of the future (shown in purple).}
    \label{fig: qualitative_plot}
\end{figure}

\textbf{Qualitative study}. Fig. \ref{fig: qualitative_plot} shows a visualization of how the agent imagines the future at a particular point while driving in a crowded intersection. In general, we observe accurate trajectories that closely resemble the human ones. Further results and failure cases are discussed in the suppl. materials.

%% file: sections/5_conclusion.tex
\section{Conclusion}
\label{sec: conclusion}

We have described DSS, a novel test-time planning framework for autonomous driving. In it, the ego-vehicle imagines how other agents will behave in the future and optimizes its own actions accordingly. It is implemented using differentiable simulation for the environment dynamics and learnable classifiers to approximate the non-differentiable collision and offroad computations. We have evaluated the framework in both tracking and autonomous path planning settings, showing strong gains compared to relevant methods. Our approach achieves better displacement rates, indicating more humanlike driving, and ensures that executed actions are selected using test-time optimization, rather than directly from a policy that is only expected to generalize. Potential future work includes transferring this idea to other simulators, additional dynamics models, and real-world situations.